\documentclass[a4paper]{svproc}
\usepackage{url,lineno}
\usepackage{pstool}
\usepackage{hyperref}
\usepackage[onehalfspacing]{setspace}
\usepackage{listings}
\usepackage{xcolor}
\usepackage{graphicx}
\usepackage{hyperref}
\usepackage{xpatch}
\usepackage{natbib}

\begin{document}
\mainmatter              
\title{A Framework for Neurosymbolic Robot Action Planning using Large Language Models}
\titlerunning{A Framework for Neurosymbolic Robot Action Planning using LLM}
\author{Alessio Capitanelli \and Fulvio Mastrogiovanni}
\authorrunning{Capitanelli and Mastrogiovanni}
\tocauthor{Alessio Capitanelli and Fulvio Mastrogiovanni}
\institute{Department of Informatics, Bioengineering, Robotics, and Systems Engineering, University of Genoa, Via Opera Pia 13, 16145, Genoa, Italy\\
\email{alessio.capitanelli@dibris.unige.it}, \email{fulvio.mastrogiovanni@unige.it}}

\maketitle

\begin{abstract}

Symbolic task planning is a widely used approach to enforce robot autonomy due to its ease of understanding and deployment in engineered robot architectures. 
However, techniques for symbolic task planning are difficult to scale in real-world, highly dynamic, human-robot collaboration scenarios because of the poor performance in planning domains where action effects may not be immediate, or when frequent re-planning is needed due to changed circumstances in the robot workspace. 
The validity of plans in the long term, plan length, and planning time could hinder the robot's efficiency and negatively affect the overall human-robot interaction's fluency. 
We present a framework, which we refer to as Teriyaki, specifically aimed at bridging the gap between symbolic task planning and machine learning approaches.
The rationale is training Large Language Models (LLMs), namely GPT-3, into a neurosymbolic task planner compatible with the Planning Domain Definition Language (PDDL), and then leveraging its generative capabilities to overcome a number of limitations inherent to symbolic task planners. 
Potential benefits include 
(i) a better scalability in so far as the planning domain complexity increases, since LLMs' response time linearly scales with the combined length of the input and the output, instead of super-linearly as in the case of symbolic task planners, and 
(ii) the ability to synthesize a plan action-by-action instead of end-to-end, and to make each action available for execution as soon as it is generated instead of waiting for the whole plan to be available, which in turn enables concurrent planning and execution. 
In the past year, significant efforts have been devoted by the research community to evaluate the overall cognitive capabilities of LLMs, with alternate successes. 
Instead, with Teriyaki we aim to providing an overall planning performance comparable to traditional planners in specific planning domains, while leveraging LLMs capabilities in other metrics, specifically those related to their short- and mid-term generative capabilities, which are used to build a look-ahead predictive planning model. 
Preliminary results in selected domains show that our method can: 
(i) solve 95.5\% of problems in a test data set of 1000 samples; 
(ii) produce plans up to 13.5\% shorter than a traditional symbolic planner; 
(iii) reduce average overall waiting times for a plan availability by up to 61.4\%. 

\end{abstract}

\section{Introduction}

Nowadays, we are witnessing the emergence of scenarios where robots are expected to operate with an increased autonomy, versatility, and robustness, especially in tasks to be carried out jointly, or in close coordination with humans.
In fact, in human-robot collaboration tasks, humans could act as simple \textit{beneficiaries} of robot behaviors, or they could act as \textit{teammates}, actually partaking in robot actions as they unfold.

This state of affairs not only holds in assistive and service robotics, but is becoming particularly relevant in settings inspired by the Industry 4.0 paradigm, which is believed to profoundly shape the nature of next-generation manufacturing operations \citep{heyer2010human}. 
These scenarios require robots to exhibit quite dynamic and reliable capabilities in sustaining a purposeful collaboration with human operators: 
while robots may be characterized by a continuum range in autonomy, also depending on the specific operation at hand, human operators may supervise robot actions and intervene if and when necessary. 
Therefore, robots must not only operate reliably in line with the interaction with human operators, but also to face unforeseen events very likely to happen when the work to be done is poorly structured, and human loosely predictable behavior is involved \citep{Darvishetal2018}. 
This paradigm shift requires robots to validate or re-plan their future actions continuously, ideally to predict or at least to react to changing conditions or the outcomes of unforeseen human behavior.
As a consequence, it turns out that planning performance is of critical importance to ensure both an efficient robot operation and an effective human-robot collaboration. 

In scenarios characterized by such requirements, conventional, symbolic task planners have limitations that hinder their efficacy. 
In current practice, planners use search algorithms to explore the range of possible states that the modeled environment can assume until they find a sequence of planning operators (that is, actions) in-between states leading to a state compatible with the desired goal situation.
In its simplest form, planning engines assume that states and actions are represented in terms of predicates, that is, \textit{fragments} modeling certain aspects of reality, and these in terms of (possibly grounded and typed) variables. 
If the \textit{closed-world} assumption is posed, that is, the truth value associated with predicates is assumed to be \textit{false} if not stated otherwise, the specific predicates part of each intermediate state depend only on the effects of the actions that generated those state.
Therefore, the planning process can be seen a generative Markov decision process based on a symbolic substrate made up of (grounded and typed) variables, predicates, and their arrangement in actions.
On average, search algorithms tend to exhibit combinatorial explosion in both time and space complexity as the number of symbols involved in the planning process increases \citep{Garrettetal2020}.
Heuristics are aimed at mitigating such unfortunate computational complexity by trading it with optimal performance and algorithmic soundness.
In spite of all the available heuristics, which often make the planning process of a \textit{satisfycing} nature, the overall computational complexity makes frequent re-planning extremely costly in terms of needed planning time, specifically considering the fact that a plan synthesized by a symbolic task planner is an \textit{atomic} entity, that is, it is either generated in its entirety, or it is not available.
Although approaches for hierarchical task planning have been proposed in the literature, whereby high-level (generic) plans may be created quickly, and low-level (specific) sub-plans can be generated in chunks, the basic issue remains, and it is also necessary to carefully engineer which part of the plan could be synthesized at the high-level, and which part (in terms of actions actually executable by a robot) should be made available in sub-plans \citep{Mastrogiovannietal2004, Muralietal2020, Darvishetal2021}.

When human operators and robots collaborate on a task, they can achieve a high standard of coordination, leading to a synchronized blending of their actions with precise and efficient timing \citep{Carfietal2019}. 
This feature is referred to as \textit{fluency} as per \cite{hoffman2019evaluating}, and it has been demonstrated that the delay experienced by a human immediately after completing an action, as incurred by their teammate, has a strong correlation with subjective fluency perception.
Obviously enough, the aforementioned combinatorial explosion of planning times negatively affects the quality of human-robot collaboration, and specifically its fluency.
It has been argued that fluency in human-robot collaboration could be improved by introducing intuitive communication channels able to make clear which action a robot will carry out next \cite{Maccioetal2022}.
In fact, such approaches work while the cooperation unfolds, whereas in case of re-planning it would be necessary to put the collaboration process on hold.

In this paper, we introduce a framework, which we call Teriyaki\footnote{Teriyaki: \url{https://github.com/alessiocpt/teriyaki}}, to train and invoke Large Language Models (LLMs) \citep{hatcher2018survey} to behave as task planners. 
Teriyaki is based on two conceptual tenets, which are reflected in the overall architectural design as well as for what concerns its practical use.
The first is related to the fact that Teriyaki is \textit{one possible instance} of a more general procedural workflow, which is aimed at generating a data-driven, domain-specific planner, whereas the second concerns a workflow structured such that:
first, a domain-specific data set of problems and plans must be generated using an existing PDDL-based planner; 
then, such data set must be used to fine-tune an LLM of choice; 
finally, the synthesized planner can be used in place of the original, traditional planner used for the generation of the data set. The substitution of the planner is trivial in any traditional \textit{sense-plan-act} architecture, but it is also possible to adapt such architecture in order to exploit the main advantage of Teriyaki planners over traditional ones, i.e., the ability to generate a plan action by action, and thus plan and execute in parallel. 

LLMs such as OpenAI's GPT-3 \citep{brown2020language} are characterized by the possibility of 
(i) linearly scaling in computational complexity with the total length of prompt and completion, and 
(ii) generating partial results while iteratively predicting the next set of symbols in a sequence.
We argue that these two features could be leveraged to design a task planner with a reduced run-time complexity, and capable of making the next action available ahead of full plan synthesis, which would allow a robot architecture embedding such planner to begin action execution before the entire plan is generated, that is, a very desirable requirement in human-robot collaboration scenarios.
If the robot architecture allowed it, this may unlock concurrent plan generation and execution. 
Moreover, if such a model could be trained to receive input and return output using the Planning Domain Definition Language (PDDL) \citep{aeronautiques1998pddl}, it would maintain full compatibility with existing software frameworks for symbolic task planning already widely adopted by the robotics community, such as ROSPlan \citep{cashmore2015rosplan}. 
We refer to the workflow underlying Teriyaki as \textit{neurosymbolic}, as it combines neural network based learning as \textit{substrate} with symbolic knowledge representation and reasoning as \textit{logic} \citep{garcez2020neurosymbolic}. 
Unlike traditional symbolic task planners, our approach is domain-dependent and requires training, but it is trivial to generate large data sets for training on a given domain using \textit{traditional} task planners. 

Teriyaki has been evaluated and tested on two planning domains, specifically for the collaborative manipulation of articulated objects by human operators and robots, a challenging task previously described in the literature \citep{capitanelli2018manipulation, bertolucci2019automated, bertolucci2021manipulation}.
These domains have been selected because of the challenges they pose on the planning process: 
(i) for symbolic task planners, because the manipulation of an articulated object scales very poorly with the number of its links, joints, and the allowed angle configurations; 
(ii) for LLMs, since the domains include \textit{conditional effects}, that is, an action could imply modifying the state of the representation in ways that are not immediately apparent from the arguments of the action itself, for example, if a link were rotated around a joint, all downward links in the chain should be also \textit{implicitly} rotated with respect to an absolute reference frame. 

The training process has been performed on a large data set of 9000 \textit{problem-plan} pairs generated automatically and solved by an existing, state-of-the-art, traditional, symbolic, PDDL-based planner, namely Probe \citep{lipovetzky2011searching}.
It is noteworthy that the data set, as well as the code to generate more pairs, are available on the accompanying repository.
The resulting models have been rigorously tested on 1000 pairs not previously used for training, and evaluated in terms of percentage of valid plans, plan length, and planning times. 
During training, data related to planning validity, defined as the percentage of plans that can be formally proved to be correct, have been collected to investigate their evolution with the growing number of training samples and to assess whether transfer learning between similar planning domains is possible.

Results show near identical planning validity between Probe and the solvers generated with Teriyaki.
However, Teriyaki outperforms Probe in terms of shorter plan length by up to 13.5\% in one of the domains. 
Regarding planning times, it is noteworthy that an objective comparison is not possible due to the different computing architectures on which the two planners run, that is, Teriyaki is based on a proprietary cloud-based service, namely GPT-3, whose response time was not guaranteed nor predictable during test sessions, whereas Probe can run on a fully accessible and controlled workstation.
In fact, in the current experimental architecture, Probe remains faster in generating a complete plan, that is, in a situation where we require Teriyaki to synthesize a full PDDL-compatible plan end-to-end.
However, when the planner as a module is integrated into a robot architecture, the use of a traditional planner like Probe forces us to adopt a pure sense-plan-act approach, that is, the plan must be synthesized in its entirety before its first action can be executed, whereas Teriyaki generates and makes available for execution the first action as it is produced. 
In our experiments, this reduces the overall waiting time by 61.4\% on average, as it exploits simultaneous planning and execution. 
Finally, we also experimentally observe that Teriyaki solvers scale with the input and output length rather than with the problem and domain complexity, hinting that there might exist more complex domains where Teriyaki could outperform Probe, and in general a traditional symbolic planner.

To summarize, the main contributions of this paper are the following: 
(i) we designed, developed, and released to the community Teriyaki, a workflow to train and use GPT-3 to solve PDDL-compatible problems for a given domain, as well as original data sets and ancillary code; 
(ii) compared to similar methods in literature, most notable the Plansformer approach \citep{pallagani2023plansformer}, Teriyaki has been tested on plans 80\% longer and including conditional effects; 
(iii) we demonstrate that LLMs can outperform non-optimal planners in terms of plan length, such as for instance Probe \citep{lipovetzky2011searching}; 
(iv) we propose to exploit LLMs capability to stream the output plan while it is being generated, for simultaneous planning and execution, leading to a significantly shorter waiting time for execution to begin, and therefore an improved fluency in human-robot collaboration.

Despite the positive results, we want to highlight that solvers generated with Teriyaki should not be considered at the moment as a complete alternative to traditional, symbolic planners.
Instead, we recommend considering them as a proof-of-concept of the planning capabilities of LLMs and their possible applications as the technology matures, especially regarding to human-robot collaboration scenarios. 

\section{Related Work}

Neurosymbolic approaches foresee that 
(i) their knowledge is encoded in vector-based representations supporting neural networks which maximize an efficient learning from data, and 
(ii) discrete symbols \textit{become available} as a result of querying and extracting knowledge from the trained network \citep{garcez2020neurosymbolic}. 
For a long time, it has been theorized that such approaches may provide an alternative to the problem of combinatorial explosion in reasoning by leveraging the reasoning mechanisms induced by the learning process.
However, it is only recently that they started to gain traction, mainly for Natural Language Processing \citep{dale2021gpt}. 
Most notably, GPT-3 and its successors GPT-3.5 and GPT-4, are famous LLMs released by the company OpenAI\footnote{Web: \url{https://openai.com/blog/gpt-3-apps}} that achieved astonishing results in generating human-level complex text in the form of structured sentences \citep{brown2020language}. Other popular models are LLama 2 \citep{touvron2023llama}, LaMDA \citep{thoppilan2022lamda}, PALM \citep{chowdhery2022palm}, Megatron-Turing NLG \citep{smith2022using} or BLOOM \citep{scao2022bloom}. 
A few of the most notable applications of these models are text summary generation and completion, sentiment analysis, and, of particular interest to our application, code generation \citep{chen2021evaluating}. 

LLMs are one of the most promising applications of \textit{Deep Generative Models} \citep{oussidi2018deep}, that is, a category of unsupervised Deep Learning algorithms capable of capturing the probabilistic distribution that can underlie the generation of a class of data.
Once estimated, it is possible to generate synthetic data compatible with such probabilistic distribution.
LLMs in particular are capable of estimating the probability of a \textit{sentence} represented as the product of each discrete symbol's probability given the symbols preceding it.
For instance, given a few words as a prompt, they can easily suggest the most likely way to complete the sentence. 
Usually, such symbols are referred to as \textit{tokens} and represent short sequences of characters.

Common ways to implement LLMs are Recurrent Neural Networks \citep{mikolov2011extensions}, Long Short-Term Memory (LSTM) networks \citep{sundermeyer2012lstm} and, most recently, Transformers \citep{vaswani2017attention}, of which GPT models are the most notable example. 
Transformers are models based on an encoder-decoder architecture, and adopt a \textit{self-attention} mechanism that allow them to weigh each part of the input data differently. 
While in the current state of affairs there are no guarantees about the long-term logical coherence of the answers given by a Transformer-based architecture, especially for long sequences of symbols, such models exhibit surprising capabilities in generating plausible and coherent lists of instructions, from cooking recipes to functioning code, surpassing by far recurrent and LSTM networks.   

Since their popularization, the integration of Large Language Models (LLMs) into robotics applications has gained significant attention, mainly as an instrument to improve the robot reasoning and planning capabilities. PaLM-E \citep{driess2023palm} explores embodied language models that incorporate real-world sensor data into language models, providing grounding for robotics problems. This approach, which includes multimodal inputs for training, shows promise in tasks such as sequential robotic manipulation planning, aligning with our aim to improve robot autonomy in dynamic scenarios. Similarly, GENOME \citep{chen2023genome} introduces a model that leverages LLMs for visual reasoning, focusing on module generation and reuse for fast task generalization, a concept that complements the goals of our Teriyaki framework.

Such interest is shared by the wider AI community, with a major focus on assessing the true extent of LLMs' general logical capabilities, especially in zero-shot to few-shot attempts. 
As per \cite{brown2020language}, for zero-shot approaches, we refer to the case when questions implying a certain level of logic capabilities are posed directly to the model, whereas one-shot and few-shot approaches provide a very limited amount of examples to the model as part of the request.
In all such cases, the model is just prompted, without further training with specific examples, a procedure now commonly referred to as \textit{fine-tuning}. 
The work by \cite{valmeekam2022large} proposes a benchmark for LLM-based planning and reasoning in a few-shot scenario. 
The authors employed PDDL, but only to automatically generate, in natural language, several logical problems never seen before by the model, and compared the performance of popular LLMs under several metrics. 
The best performer could solve only $25$ out of $500$ problems, that is, 5\%. 
Better results were obtained by \cite{wang2023grammar} using few-shot learning and \textit{grammar prompting}. 
In this approach, GPT-3.5 and GPT-4 are used to predict a specialized grammar given a test input, and then generate the output according to the rules of the grammar itself. 
Among others, the authors tested inputs in PDDL format extracted from popular but rather simple PDDL domains. 
While the benefits of grammar prompting are evident, in absolute terms they obtained mixed results depending on the domain, with success rates ranging from 40\% to 100\%. 
LLM-Planner \citep{song2023llm} is another few-shot method based on GPT-3 and designed specifically for embodied reasoning, that is, techniques aimed at being used in agents dealing with real-world, physical environments. 
This approach uses near-natural language and is characterized by a dynamic re-planning mechanism grounded on what an agent could observe in the environment at each step. 
High-level planning validity in a variety of tasks ranges from 26\% to 51\%. 
Similar approaches are discussed also by \cite{logeswaran2022few} and \cite{wake2023chatgpt}.
\cite{singh2023progprompt} proposed instead Progprompt, a method that exploits the strong performance of GPT-3 in code generation by prompting the model to generate executable Python code to solve a given problem. 
\cite{silver2023generalized} further improved on this concept by proposing a method for generalized planning in PDDL domains by also including an automated debugging mechanism. 

As perfectly stated by \cite{huang2022language}, the main issue with almost all these approaches is that plans produced naively by LLMs cannot map precisely to admissible actions, thus most of them attempt to increase their success rate in achieving the desired goal state by clever prompting approaches. 
\cite{silver2022pddl} face similar challenges while trying to solve actual PDDL problems, maintaining the PDDL format both in input and output. 
As an alternative solution, they propose to use the good-but-imperfect LLM-generated output to support a traditional (that is, based on heuristic search), PDDL-based planner. 
While this approach mitigates the issues described before and often provides planning time improvements above 10\%, the ability to generate a plan action-by-action in a way compatible with human-robot collaboration scenarios is lost.

Hence, it appears that with currently available LLMs, solving task planning problems reliably, that is, with a success rate comparable to that of traditional, PDDL-based planners, is only possible through an appropriate \textit{fine-tuning} process. 
Such an approach involves the further training of a base model with specific examples, namely problem-plan pairs, related to the planning domain at hand. 
Plansformer is an approach of this kind proposed by \cite{pallagani2023plansformer}. 
The authors trained five different models by fine-tuning CodeT5 \citep{wang2021codet5}, an LLM specialized in code generation, to solve just as many simple, well-known, PDDL-based domains such as the \textit{Tower of Hanoi}, where Plansformer reached 97\% valid plans. 
The fine-tuning dataset has been generated by solving 18000 randomly generated problems with the Fast-Downward planner \citep{helmert2006fast}, and then translating both problems and plans into a more compact form. 
PDDL-based problems are augmented by listing the actions in the domain to explicitly teach the model about their preconditions and effects.
While Plansformer demonstrates the possibility of adopting a fine-tuning approach to train solvers for different domains, the considered planning domains are rather simple, and do not include such advanced planning concepts as conditional effects, for example. 
In a more general sense, a possible interpretation of these results could be that there are obvious similarities between the capability of LLMs to learn the probability of a symbol following the preceding ones, and the search heuristics that power traditional PDDL-based planners. 
Hence, we are not just extracting knowledge from the model and exploiting its general reasoning capabilities, but rather we are training the model to approximate the search heuristic of the planner generating the dataset.

Teriyaki 
takes a similar approach with a few key differences: 
(i) we assume that LLMs are capable of handling unaltered PDDL-based problem and plan descriptions, which in turn makes the model easier to use in existing robot architectures; 
(ii) with the aim of reducing the length of the prompt, which is one of the main LLMs' constraints, we instead exclude some information, on the assumption that the model can learn it and, for example, no explicit knowledge of the action preconditions and effects is given to Teriyaki during fine-tuning;
(iii) it is trained on a dataset of non-optimal plans, which allowed us to observe that LLMs are capable of outperforming the traditional planner they are trained upon in terms of average plan length in selected domains;
(iv) we purposely focus on complex planning domains, that is, domains which require advanced planning constructs to generate provably valid plans, such as conditional actions; 
(v) we propose to use Teriyaki to \textit{stream} plans as they are generated, that is, action-by-action, to enable simultaneous planning and execution in robot tasks, specifically in human-robot collaboration.

\begin{figure}[t!]
\centering
\includegraphics[width=\columnwidth]{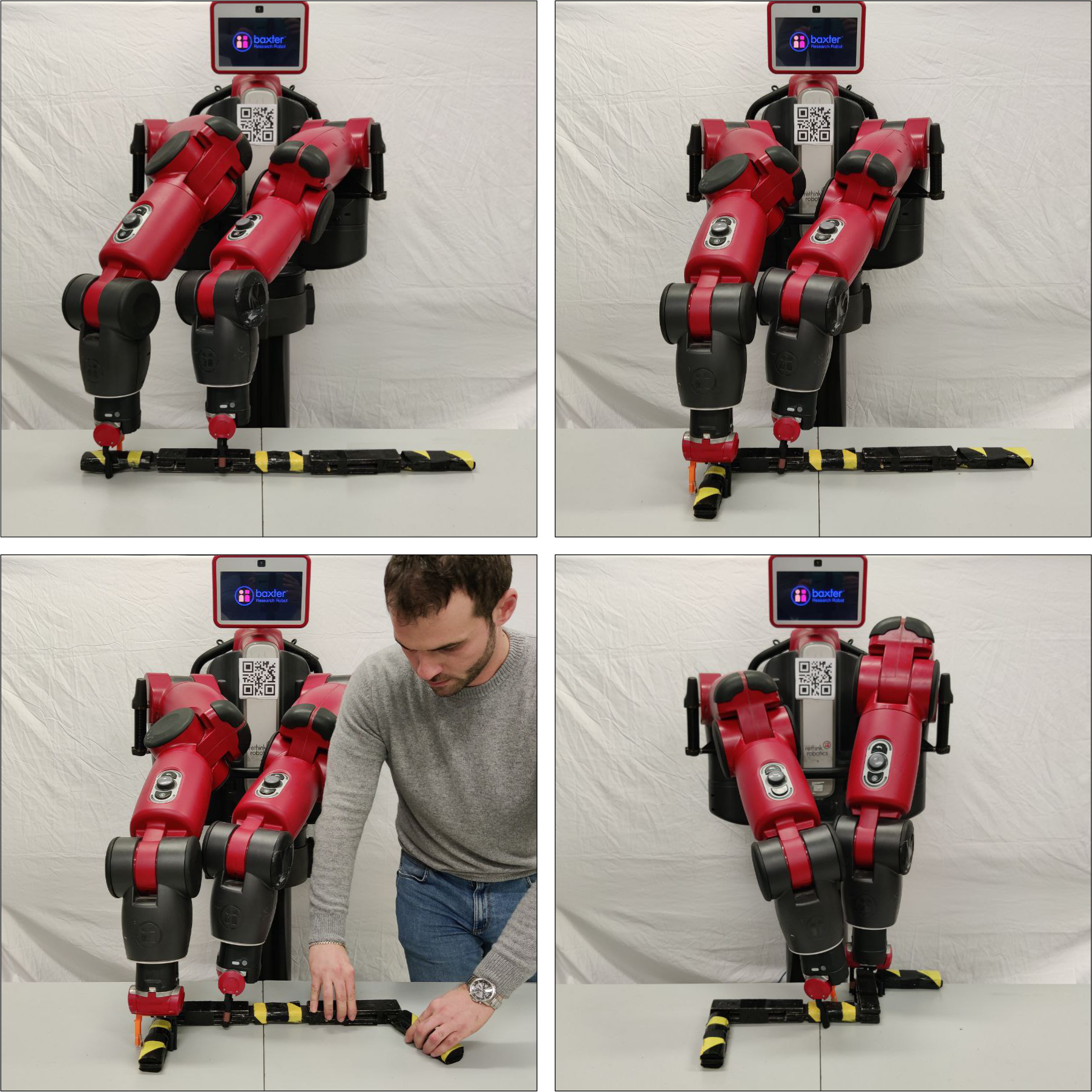}
\caption{
A Baxter robot executing actions in two domains involving the manipulation of an articulated object. 
A human can act on an articulated object's joint at any time, forcing the robot to re-plan.}
\label{fig:robot}
\end{figure}

\section{Methods}

\subsection{Planning Domains, PDDL Version, and Planner}

In our work, and with the aim of challenging both the chosen, traditional, PDDL-based planner as well as Teriyaki, we have selected two PDDL domains, which we refer to from now on as MACRO and NO-MACRO.
The two domains model manipulation actions on an articulated object, and have been previously showcased in human-robot collaboration scenarios.
It is noteworthy that, for the work described here, we have used the same robot perception and control architecture that has been discussed by \cite{capitanelli2018manipulation}.
Such an architecture includes vision-based perception modules able to correctly detect and identify each link of the articulated object, as well as their pose; the outcome of these modules is a series of symbolic predicates which specify the overall object configuration for planning and monitoring purposes.
Likewise, robot control aspects related to manipulation are included in the architecture, and suitable mappings between PDDL-like, symbolic actions and robot motions are available. 
Figure \ref{fig:robot} shows a Baxter robot executing a plan from the two domains, and a possible interaction with a human.
Both domains use the so-called \textit{absolute} representation as described by \cite{capitanelli2018manipulation}, meaning that angles between pairwise links of the articulated object are expressed with respect to an absolute reference frame. 
The choice is motivated by the fact that domains using an absolute representation require \textit{conditional effects} in the planning domain, that is, actions can have implicit effects on joint angles not directly affected by the outcomes of the action itself, and therefore the planner is expected to be quite stressed in maintaining all the constraints.
In fact, if managed by a traditional, symbolic action planner this requires propagating the (conditional) effects of each action to modify the value of state variables not directly affected by the effects.
Likewise, we argue that in the case of LLMs like GPT-3, conditional effects could stress the model because it should be harder to maintain coherence in the plans given the underlying generative process.

The main difference between the two selected domains is that one uses PDDL \textit{macros} whereas the second does not. 
Macros are ordered, compound actions that bundle together a number of basic actions, for example, a \verb|grasp-rotate-release| action instead of three \textit{atomic} ones, that is, \verb|grasp|, \verb|rotate|, and \verb|release|.
Their use is an effective way to reduce the planning time in traditional action planners at the cost of accepting \textit{less optimal} plans.
In fact, the use of macros could lead a planner to enforce the use of an actions sequence, that is, a given macro, at the cost of introducing possibly spurious actions. 
For example, if the goal state assumed the articulated object to be in a given configuration, but with the robot gripper still maintaining its hold on the object, the planner could prefer using a macro \verb|grasp-rotate-release| followed by another \verb|grasp| action (partially compensating the effects of the macro) instead of two atomic actions \verb|grasp| and \verb|rotate|.
Macros are also supposed to facilitate the generative process of GPT-3 since the use of macros is expected to shorten the resulting plan significantly. 

Since the two domains are fundamentally similar but lead to plans of different lengths, they are ideal candidates to test how Teriyaki-based solvers scale with the output length. 
As both domains taken into consideration here used conditional effects, we used PDDL 2.1 and a compatible PDDL-based planner, namely Probe \citep{lipovetzky2011searching}, one of the planners used in the previous work by \cite{capitanelli2018manipulation}.
Probe is a satisfycing planner which heuristically tries to generate single action sequences by exploring the search space using a greedy, depth-first approach (that is, a \textit{probe}).
Statistically, a solution can be quickly found in this way, albeit not necessarily the optimal one.
Probes can be used as first solutions on top of which it is possible to find better (shorter) plans.
This could be done by analyzing probes in order to find landmarks, that is, certain states that are reached by different probes, and therefore represent important states to generate.
If landmarks are available, it is then possible to obtain sub-plans in-between landmark states.
Probe has been selected for two reasons.
On the one hand, its use in previous work of ours allow us to make a direct comparison between its outcomes and plans generated by Teriyaki; on the other hand, its widespread use could allow for an easier adaptation of our work in different scenarios.
In fact, the choice of Probe has an impact on the quality of the results, and it should pointed out that better alternatives could be adopted.

\subsection{Choice of the LLM, Data Set Generation and Composition}
\label{dataset}
\begin{figure}[t!]
\centering
\includegraphics[width=\columnwidth]{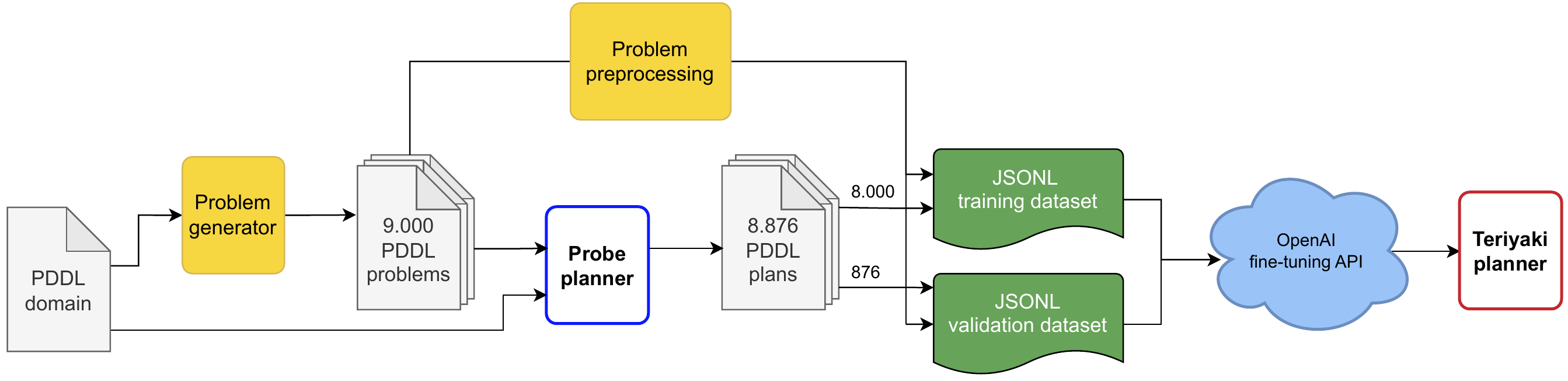}
\caption{
A diagram of Teriyaki fine-tuning process.
Blocks in yellow represent custom code developed for data generation and processing as described in Section \ref{dataset}.}
\label{fig:training}
\end{figure}

As we anticipated, in this work we leverage GPT-3, a closed-source model and commercial product that at the time of this writing can only be accessed as a cloud-based service through OpenAI's API.
Our results are not limited to the specific features of GPT-3 and could in principle be replicated with any available LLM. 
In practice, since our first experiments with GPT-3 presented in this paper, we have tested several smaller models that can be run locally, such as LLama 2 7B, 13B, and 30B, but none of them could reliably solve problems in the two selected domains, and therefore the related results are not reported here. 
Forced to employ a commercial model as a service, the choice of GPT-3 is motivated by its size and speed compared to its more recent versions.
In the case of GPT-3, the training can be performed by providing a structured file where each line is a \textit{prompt-completion} pair. 
We assumed that the GPT-3 model can implicitly learn by example the rules usually defined in a PDDL domain file, such as the allowed (modeled) actions, and their respective preconditions and effects. 
Hence, we conducted training sessions using only \textit{problem-plan} pairs, whereas we use the problem as a prompt and the plan as completion. 
We remark here that only part of the PDDL problem is used as a prompt, as many of the predicates are actually \textit{non functional}, that is, they are used to encode general properties of the problem, such as the relative order of joint angle configurations, an example being: a joint at $45$ \textit{deg} can be rotated clockwise to $60$ \textit{deg}, and counter-clockwise to $30$ \textit{deg} in $15$ \textit{deg} increments. 
These predicates remain the same in all problems and are never modified by action effects.
Therefore, we assume that the model can learn them from the implicit information encoded in the plans, in the same way as the domain. 
We thus removed them from the prompt to reduce its length, as prompt length has an impact on response time and the total prompt plus completion length of each query cannot exceed $2048$ tokens, or \textit{circa} $8000$ characters, for the selected model, allowing us to generate longer plans without truncation and improving performance.

At the time the experiments were performed, OpenAI's documentation suggested using the largest and most powerful version of the model, called \verb|davinci|, for a conditional generation, that is, generating original text based on input. 
For this use case, the documentation recommends training the model using at least $500$ training samples and aiming for $8000$.
One of the advantages of Teriyaki is that we can easily generate large training datasets using randomly generated problems and a PDDL-based action planner.
Therefore, we generated $9000$ problems, out of which $8000$ were reserved for training and $1000$ for validation.
It must be noted that out of the $9000$ planning attempts performed by the planner $124$ failed, so the validation set has been reduced to $876$ samples. 
Finally, we added another $1000$ problem-plan pairs as a test set, verified that each sample within this test set was solvable by Probe, and guaranteed that no problem instance was a duplicate of those existing in the training and validation sets, thereby preventing any cross-contamination between the datasets.

The next step requires validating the completeness of the plans generated by Probe to ensure that we train our model only on correct data. 
To do so we use the widely known VAL validation suite \citep{howey2004val}, which requires plans to be formatted in the International Planning Competition (IPC) standard\footnote{Web: https://ipc2023.github.io/}. 
As Probe does not return plans compliant with this standard, we developed custom code to reformat the plan files to be compatible with VAL. 
This is also necessary to ensure that Teriyaki will be trained on IPC-compatible standard plans, and will thus later reply to our queries in the same way, allowing us to easily benchmark its validity as far as planning is concerned. 
Running VAL over the dataset resulted in all plans passing the test, meaning that they reached the desired goal state, even though it must be reminded that Probe previously failed to generate a plan \textit{at all} for $124$ out of $9000$ problems, that is, about $1.37\%$ of the total.

Finally, we compiled the training and validation datasets in \verb|jsonl| format. 
Each line of a dataset file is composed of a \textit{prompt}, that is, the problem, and a \textit{completion}, that is, the plan. 
We added to the end of each prompt and completion the respective termination sequences, namely \verb|\n\n###\n\n| and \verb|END|. 
As the name suggests, the termination sequences signal to the network the end of the message. 
This is especially important for the completion as the termination sequence can then be used as a stopping condition when querying the model. 
A sample line is provided in Listing \ref{lst:jsonl}, edited for the sake of brevity and clarity. 
The whole data generation and dataset preparation process is outlined in Figure \ref{fig:training}.

It is important to clarify that the generated data utilized in our research should not be perceived as synthetic with the negative connotation of \textit{less real}. Given that our focus is on symbolic task planning and assuming flawless perception and grounding by the robot, this data is indistinguishable from real data from the perspective of the planner. Consequently, for the purposes of this work, we exclusively employ generated data for training and analysis, as it provides equivalent validation for the planning module without introducing variances and biases from perception and execution modules.

\vspace{11pt}

\colorlet{punct}{red!60!black}
\definecolor{background}{HTML}{EEEEEE}
\definecolor{delim}{RGB}{20,105,176}
\colorlet{numb}{magenta!60!black}

\lstdefinelanguage{json}{
    basicstyle=\small\ttfamily,
    numberstyle=\scriptsize,
    stepnumber=1,
    numbersep=8pt,
    showstringspaces=false,
    breaklines=true,
    frame=lines,
    backgroundcolor=\color{background},
    literate=
     *{,}{{{\color{punct}{,}}}}{1}
      {\{}{{{\color{delim}{\{}}}}{1}
      {\}}{{{\color{delim}{\}}}}}{1}
      {[}{{{\color{delim}{[}}}}{1}
      {]}{{{\color{delim}{]}}}}{1},
}

\lstset{frame=tb,
  language=json,
  firstnumber=1,
  label={lst:jsonl},
  caption={An example of training sample.},
  captionpos=b}

\begin{minipage}{0.95\linewidth}
\begin{lstlisting}
{"prompt":
"(:init (angle_joint angle315 joint1)
(angle_joint angle300 joint2)
(angle_joint angle285 joint3)
(in-centre joint2) 
(free gleft) (free gright))
(:goal (and
(angle_joint angle0 joint1)
(angle_joint angle300 joint2)
(angle_joint angle285 joint3)))
\n\n###\n\n", 
"completion":
" 0.00100: (link-to-central-grasp ...)
0.00300: (increase_angle_first_child_45 ...)
0.00500: (release-links ...)
0.00700: (link-to-central-grasp ...)
0.00900: (decrease_angle_first_child_45 ...)
END"}
\end{lstlisting}
\end{minipage}

\subsection{Training}
\label{finet}

As far as the training process is concerned, we decided to run a complete 8000-sample training on the MACRO domain, since plans in the related dataset are shorter on average, and therefore \textit{supposedly} easier to be learned by GPT-3. 
We call this model Teriyaki-MACRO.

When we train an LLM to perform action planning, the cost function associated with the resulting generation process rewards linguistic coherence rather than planning validity.
This means that we are hypothesizing \textit{linguistic coherence} as a proxy for \textit{logical coherence}, and therefore we must assume that validation during the training process differs from the planning validity of the resulting model, whereas ``planning validity'' is defined as the percentage of plans that are 
(i) formally correct and 
(ii) able to reach the desired goal state, or a state compatible with it. 
By ``formally correct'' we mean that each action's arguments are consistent both with the specifications given in the planning domain and the state of the modeled reality when that action is carried out, that is, the action preconditions are not violated.
To be able to measure how planning validity increases with the size of the provided training set, we decided to perform the training process in steps. 
At each step, we provided samples to double the total size of the training set. 
Starting from the minimum amount of 500 samples, we then trained the system with 1000, 2000, 4000, and 8000 total samples, and saved a snapshot of the trained model at each step.

As anticipated above, the base model chosen for the training process is GPT-3 and specifically \verb|text-davinci-002|. 
Regarding the hyper-parameters, the number of training epochs was set to 2, while the batch size and the learning rate multiplier were left to their default values. 
The batch size defaults to $0.2\%$ of the number of examples in the training set, while the default learning rate is automatically determined in a range from $0.05$ to $0.2$ depending on the final batch size. 
We also highlight that since the model was effectively fine-tuned five times, the learning rate was reset to the default value at the beginning of each session, with an effect similar to gradient descent with warm restarts \citep{loshchilov2016sgdr}.
We did provide the optional validation set and enabled the \verb|compute_classification_metrics| options to obtain training statistics for further analysis. 
The total training cost for this procedure on a single planning domain at the time of training (01/12/2022) was around 250\$.

\subsection{Transfer Learning}
\label{sec:transfer}

For the NO-MACRO domain, we used the same methodology, but we decided to generate two candidate models.
The first model is trained starting from the \verb|text-davinci-002| model as before, whereas the second is trained starting from the MACRO model obtained previously. 
We call these models Teriyaki-NO-MACRO (davinci) and Teriyaki-NO-MACRO (MACRO).
The hypothesis is that as domains share many concepts and the MACRO model has already been primed for them, the second candidate should reach a higher planning validity with a smaller amount of training samples. 

Results shown in Section \ref{restrans} confirmed this hypothesis, so we interrupted training when Teriyaki-NO-MACRO (MACRO) reached comparable results to its parent model and discarded entirely Teriyaki-NO-MACRO (davinci). 
Regarding the training dataset, the only difference with the example provided in Listing \ref{lst:jsonl} is that the prompt part is preceded by the \verb|\n--NO-MACRO| tag. 
This tag was introduced to test whether the model could be used to solve problems for both the MACRO and the NO-MACRO domains, by simply adding the tag in our queries to the system. 
Unfortunately, the NO-MACRO models loses the ability to solve problems in the MACRO domain, suggesting that to generate models that can solve multiple domains, \textit{training should be performed including examples from all the domains of interest}. 

\subsection{Invoking the LLM and Streaming the Plan}
\label{usgt}

After the training phase, it is possible to query the model through an API call by providing as a prompt the PDDL predicates describing the initial and goal states for the robot and the articulated object, as one would do with a traditional, symbolic PDDL-compatible action planner. 
Several parameters can be configured, and below is a list of those that can impact the overall quality of the resulting plan.
\begin{itemize}
\item \verb|temperature| is the most important as it controls how \textit{deterministic} the reply (and therefore the plan) will be. 
Keeping it at $0$ corresponds to an \verb|argmax| operation on the next token probability and ensures that we always get the most likely result. 
While in a regular text generation some level of \textit{creativity} is desirable, we strongly recommend keeping it at $0$ for plan generation, especially when robot actions are planned in the context of a human-robot collaboration scenario.
\item \verb|presence_penalty| and \verb|frequency_penalty| can assume values between $-2$ and $2$, and can be used to reward or penalize repetitions in the generated text. 
We have observed that setting the former to $2$ seems to improve planning validity by ${1-2}\%$ and \textit{vice versa}, but at the moment we have not investigated enough this effect, so we decided to set the default value to $0$ for both in our tests.
\item \verb|stop| can be used to set a string as a stopping condition, meaning that when that string is generated by the model, the generation immediately terminates. 
Coherently with our training dataset, we set this value to \verb|END|.
\item \verb|max_tokens| controls the maximum response length, thus we recommend setting it as high as possible to minimize the chance of a plan being truncated. 
Since each model has a maximum total prompt plus completion length, and the worst-case prompt length depends on the planning domain at hand, this value should be assessed case by case. 
In our case, $1900$ appears to be the highest value for robust operation.
\end{itemize}

To stream the plan we can set the parameter \verb|stream| to \verb|true|. 
In this way, the model starts returning tokens as soon as they are available, instead of waiting for the full sequence of tokens to be generated. 
This has no impact on the time needed to receive the full answer, but it does reduce significantly the time needed to receive the first action, that is, the response time.

\subsection{An Architecture for Simultaneous Planning and Execution}
\label{sec:arc_th}

\begin{figure}[t!]
\centering
\includegraphics[width=\columnwidth]{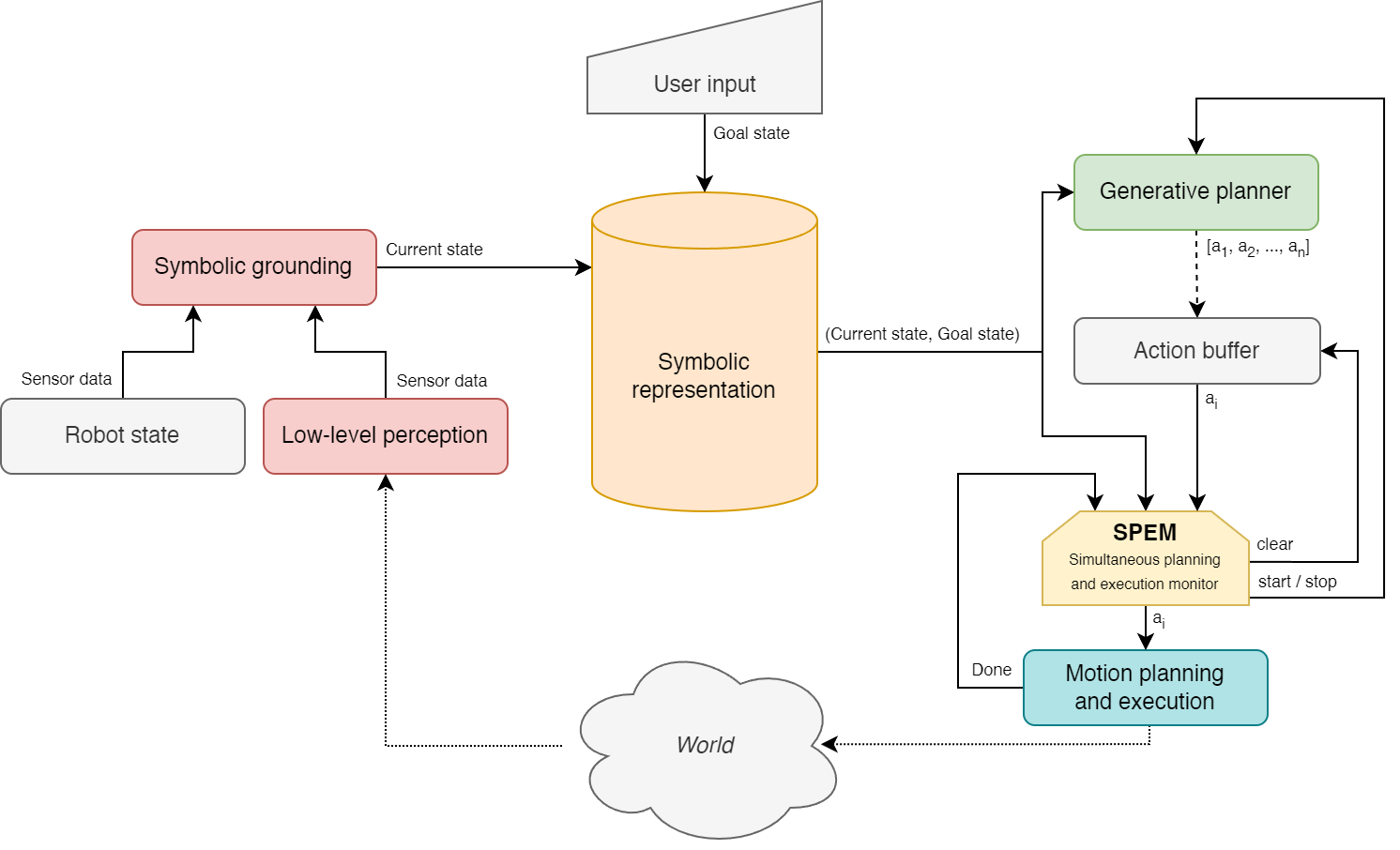}
\caption{
A possible architecture integrating a generative Teriyaki planner to implement simultaneous planning and execution. 
The dashed line below the planner represents the fact that actions are generated one by one, and then buffered waiting for their execution.
The SPEM module checks that each action's preconditions are satisfied before submitting it for execution.}
\label{fig:arch_diagram}
\end{figure}

In a traditional sense-plan-act data flow, the system works through the following steps:
(i) first collects data about the robot state and its workspace from endogenous and exogenous sensors;
(ii) grounds such data into symbols to represent the current state;
(iii) generates an end-to-end plan to reach a given goal situation from the current state;
(iv) begins to execute all the planned actions in sequence, one after the other, possibly monitoring their outcomes and halting execution in case of unforeseen disturbances, errors and, in general, any misalignment between the expected and the current state.

With Teriyaki, and building up on the work done by \citep{capitanelli2018manipulation}, it is possible to modify this simple data flow as shown in Figure \ref{fig:arch_diagram}.
While step (i) and step (ii) remain unchanged, now the data-driven synthesized planner can output an action as soon as it has been generated by the model. 
This can be achieved simply by considering as an action every new line generated by the planner.
Since the generation of the plan cannot generally be paused, actions are stored in a first-in first-out buffer. 

In order to monitor each action execution, we introduce a specific module, referred to as Simultaneous Planning and Execution Monitor (SPEM).
After each action has been executed, SPEM verifies two conditions:
(a) it checks whether the overall goals have changed during the execution of the last action, for example, due to new human actions;
(b) it considers the most up to date current state and verifies, using the VAL validation utility, that such state is compliant with the next action's preconditions.
If both conditions are met, SPEM submits the next action to motion planning and execution; otherwise, it resets the planner and clears the action buffer.
If this happens, planning can be restarted taking into consideration the most up-to-date current and goal states.

Condition  (b) is essential because the planner continues to generate a plan action-by-action based on the state available when it was first initialized.
Yet, the current state changes continuously, either in expected ways, as a result of the execution of actions, or in unexpected ways, as a result of unforeseen disturbances and human actions.
In the second case, the next action might require preconditions that the planner believes are met but in reality, are not anymore.
Therefore, SPEM both prevents catastrophic failures, and allows for seamlessly recovering from disturbances, or adapting to human behavior.
Considering that planners generated with the Teriyaki workflow can generate a plan action-by-action, this recovery procedure only takes the time needed to generate a single action. \textit{Vice versa}, in the traditional sense-plan-act data flow, the overall system would be forced to stop the execution until a complete plan was generated by a traditional planner.

Two remarks must be made.
The first is that a module like SPEM would not be necessary if we only generated a single action at a time, and restarted the planner with the same goal state but updated knowledge of the world afterward. Unfortunately, in our tests, this operation modality does not lead to reasonable plans, as the planner loses the context of previously taken actions and then easily falls into loops, such as, for example, centering the articulated object around a given joint, and then again centering it around another, and so on.
This phenomenon is conceptually similar to the well-known Sussman Anomaly \citep{sussman1973computational}, and is completely dealt with by the functionality encoded in SPEM.
The second is related to the fact that, conceptually speaking, this workflow is still a traditional sense-plan-act architecture.
Indeed, comparing this robot control architecture \textit{side by side} with the one presented by \cite{capitanelli2018manipulation}, the practical behavior of the robot would be indistinguishable, except for the reduced waiting time whenever a replanning process is needed.
For this reason, in Section \ref{sec:arch_exp} we decided not to present execution timings and instead focus on a comparison of waiting times before execution between traditional and Teriyaki-based planners.

\section{Results}

\subsection{Relation between Token and Planning Validity with an Increasing Number of Training Samples}
\label{accuracyres}

\begin{figure}[t!]
\centering
\includegraphics[width=\columnwidth]{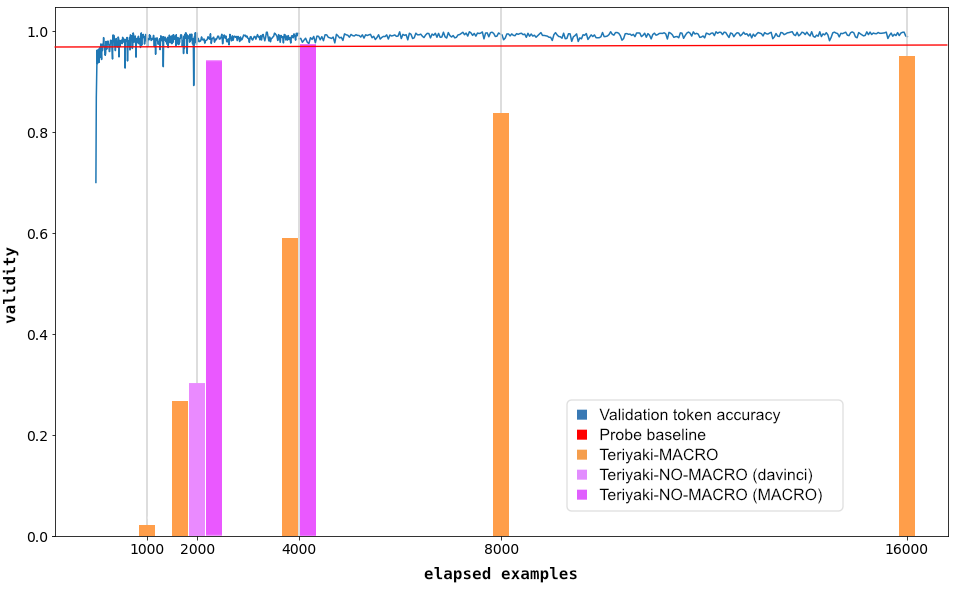}
\caption{
Evolution of the token validation accuracy and the planning validity as the number of training examples increases.  
The blue line represents the evolution of the validation token accuracy during learning as reported by GPT-3 classification metrics. 
The bars represent the planning validity of the Teriyaki-MACRO, Teriyaki-NO-MACRO (davinci) and Teriyaki-NO-MACRO (MACRO) models.
The red line represents our baseline, that is, the percentage of plans solved by Probe.}
\label{fig:validation}
\end{figure}

In Section \ref{finet}, we pointed out that we use linguistic coherence as a proxy for logical coherence. 
Figure \ref{fig:validation} compares the evolution of the validation token accuracy and the planning validity for the MACRO model, against the number of examples used to train the model itself. 
On the one hand, the validation token accuracy measures the accuracy at predicting the next token, that is, approximately the next 4 characters of the response, in our case coinciding with the plan, on the validation set.
On the other hand, we refer here to planning validity as the percentage of plans in the 876-sample validation set that are both formally correct and reach the desired goal state, that is, which one passes the VAL validation utility test. 
For the former, data are retrieved from the classification metrics reported by GPT-3 itself after training, where this information is available every 8 training steps. 
For the latter, we used the snapshots of the model taken after training with 500, 1000, 2000, 4000, and 8000 examples.
Such snapshots are used to plan against all 876 problems in the validation set in the conditions described in Section \ref{usgt}. 
VAL is run on the obtained plans and finally, the planning validity can be computed. 
In Figure \ref{fig:validation}, the evolution of planning validity is represented by the orange bars to highlight the fact that the value is measured at each snapshot. 
It must also be noted that the parameter \verb|elapsed_examples| does not correspond to the number of unique examples in the training set, but it is scaled by a factor of 2 because we used two epochs for training, thus each example was used twice. 

In Figure \ref{fig:validation} the orange bars report the validation accuracy of the Teriyaki-MACRO model. 
It can be observed that GPT-3 reaches a very high validation token accuracy after the first 500 samples, as expected by a model well-known for its few-shot learning capabilities. 
Nevertheless, planning validity rises at a much slower pace. 
Like in few-shot methods, the LLM cannot yet map precisely to admissible actions and even a single mistaken token could break a plan that would be otherwise \textit{linguistically} coherent. 
A very common mistake in plans generated by the model trained with only 500 samples is that the model ignored the conditional effects of actions.
As we correctly hypothesized, conditional effects can be quite challenging as they imply propagating their effects downstream in the plan to keep its overall semantic coherence.
In this case, actions are correct and reasonably parameterized but they do not meet the necessary preconditions as they ignore that a given joint is in a state different from the one expected by the model, due to the \textit{indirect} effects of a previously planned action. 
Eventually, the model reaches a very high $95\%$ planning validity on the validation set after training over $8000$ unique samples. 
Despite this, the model does not seem to overfit to the problems used for training, as shown by the results on the test set presented in Section \ref{sub:solvers}. Hence, it appears that a larger training data set or a higher number of epochs might result in a even higher planning validity. 

Results of planning validity are strikingly similar to those obtained by the Plansformer \citep{pallagani2023plansformer}, yet they have been achieved with a training set of 8000 instead of 18000 and 2 training epochs instead of 3.
It is worth noting that we consider the Plansformer as one possible baseline for Teriyaki, because of the conceptual connections between the two approaches, that is, the use of a fine-tuning process to adapt a foundational model with specific examples. 
It should also be noted that in spite of the similar goal, Plansformer and Teriyaki aim at demonstrating different features of the overall process: while Plansformer is focused on the fine-tuning process with the goal of demonstrating that it can be adapted to different planning domains, Teriyaki by purpose aims at demonstrating how tricky planning domains that arise in real-world robotic applications can be managed by a data-driven, LLM-based planner, namely advanced planning concepts and longer problem-plan pairs.
While our result could point to GPT-3 having stronger performance in this task than CodeT5, it is necessary to emphasize that the presence of different base models, the selection of a different PDDL-like planner, the use of different PDDL constructs, and the different prompting strategy urge an analysis well-beyond the scope of this paper. 

During this experiment, we also measured the number of planning attempts that failed because of their excessive length. 
While this number was always small compared with the validation set size, the number of failures decreased from $24$ in the first training step, down to $1$ in the final model, further suggesting that the model becomes better at generating shorter plans and avoiding unnecessary actions.

\subsection{Transfer Learning}
\label{restrans}

As previously discussed in Section \ref{sec:transfer}, the Teriyaki solver for the NO-MACRO planning domain has been chosen starting from two base candidates, namely Teriyaki-NO-MACRO (davinci) and Teriyaki-NO-MACRO (MACRO). 
As we trained the two models, we kept track of the planning validity as described in Section \ref{sec:transfer}. 
The light pink and dark pink bars in Figure \ref{fig:validation} report results in planning validity for the Teriyaki-NO-MACRO (davinci) and Teriyaki-NO-MACRO (MACRO) models, respectively. 
After 1000 samples, the Teriyaki-NO-MACRO (davinci) model reached a planning validity of $32.5\%$, while Teriyaki-NO-MACRO (MACRO) reached $95.2\%$. 
Because of this result, we immediately dropped the former model, while we further trained the latter using $2000$ samples. 
At this stage the Teriyaki-NO-MACRO model reached a validation planning validity of $98.8\%$, exceeding the $95\%$ validation planning validity achieved by the MACRO model after $8000$ samples. 
Because of this result, we decided not to proceed with further training and from now on, when we simply refer to Teriyaki-NO-MACRO, we actually mean Teriyaki-NO-MACRO (MACRO).

There are two noteworthy aspects of this experiment. 

Firstly, Teriyaki-NO-MACRO (davinci) model still reached a higher planning validity at $1000$ samples than Teriyaki-MACRO. 
This result might suggest that against our initial assumption, the NO-MACRO model is easier to learn for GPT-3 than the MACRO one. 
This result must be explored more in-depth, but it could be related to the number and quality of the actions in the planning domain.

Secondly, the transfer learning experiment within the Teriyaki framework between MACRO and NO-MACRO domains highlights its capacity for adaptation to alternate knowledge representations. However, it merits noting that both domains represent the same essential task of collaborative manipulation of articulated objects, possessing comparable action sets and state spaces. Consequently, the efficacy of transfer learning between two inherently distinct tasks also remains an area for future exploration.

\subsection{Comparison of Solvers}
\label{sub:solvers}

\begin{figure}[t!]
\centering
\includegraphics[width=0.8\columnwidth]{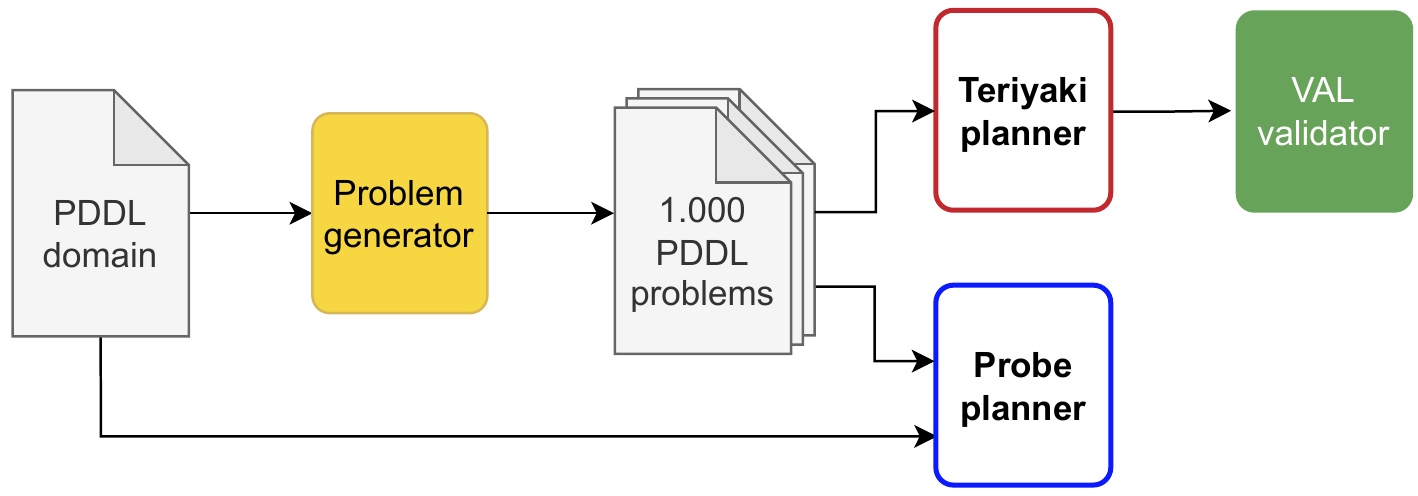}
\caption{
A diagram of Teriyaki testing process as described in Section \ref{sub:solvers}.}
\label{fig:testing}
\end{figure}

We tested the performance of both Teriyaki-MACRO and Teriyaki-NO-MACRO models in terms of planning validity, plan length, and planning times on a test set of $1000$ problem-plan pairs not previously used for training and validation in order to avoid cross-contamination, and we compared the results to the performance of the traditional action planner Probe.
The process is summarized in Figure \ref{fig:testing}.

We want to remark that, due to the significantly different computing workflows used to run Teriyaki and Probe (basically, a cloud-based architecture accessed via proprietary APIs and software installed on a local machine, respectively), a fair comparison between them in terms of planning time is not possible. 
Therefore, results about planning times in this Section are only meant at providing a baseline for future work and to show how planning times scale differently for each Teriyaki planner in the two different planning domains taken into consideration in this work.
However, Probe remains a significant baseline approach because it is the planner used to get training data for Teriyaki; although Probe itself is not an optimal a planner in terms of plan \textit{quality}, nonetheless one might realistically expect that Teriyaki's overall performance in plan generation should be related to its outcomes.   

Probe ran on an Ubuntu 20.04.4 Windows Linux Subsystem (WSL) environment, deployed on a machine equipped with an $8$-core Ryzen 3700X 3600 Mhz CPU and 16GB@3600 Mhz RAM. 
We recorded planning times as reported by the planner itself together with each generated plan. 
As far as planning validity is concerned, all plans generated were valid, but we considered the instances in which the planner failed to generate a plan as failures.

Regarding Teriyaki models, we prompted them using the settings described in Section \ref{usgt}, then verified the validity of the obtained plans using the same VAL validation tool employed at the data set generation phase. 
As the OpenAI API does not provide a method to log the exact execution time of a call, planning times of Teriyaki solvers have been measured by disabling the option to stream GPT-3 responses, and recording the time it took each API call to return a full plan to the client application. 
For this reason, it must be noted that it is impossible to discern how long each call has been queued by the cloud application before execution. 
We assessed that the effect of queuing is not negligible as running tests after the official release of ChatGPT, another popular GPT-3 derived product by OpenAI, led to longer planning times than previously recorded, possibly due to the increased traffic to the servers. 
In order to partly mitigate this effect, the tests presented here were performed during the weekend, preferably in the morning CET time. 
In Table \ref{tab:testdate} we also include the date and the starting and finishing time of each test session for reference.

\begin{table}[t!]
\caption{Summary of the tested models and date of testing}
\label{tab:testdate}
\begin{center}
\begin{tabular}{|c|c|c|c|c|c|}
\hline
\textbf{Model} & \textbf{Base model} & \textbf{Test date} & \textbf{Start} & \textbf{Finish}\\
\hline
\hline
Teriyaki-MACRO & \verb|davinci| & 18/02/2023 & 11:17:56 & 13:47:41\\
Teriyaki-NO-MACRO & MACRO & 19/02/2023 & 17:22:24 & 21:25:58\\
\hline
\end{tabular}
\end{center}
\end{table}

For all solvers and models, plan length has been computed simply as the number of actions in the plan. Table \ref{tab:results} compares the Teriyaki-MACRO and Teriyaki-NO-MACRO models against Probe in the respective domains, in terms of validity (in percentage), average plan length, maximum and average planning times, as well as the standard deviation of the planning times (in seconds). 

\begin{table}[t!]
\small
\caption{Comparison of Teriyaki models against Probe in their respective domains on the test dataset}
\label{tab:results}
\begin{center}
\begin{tabular}{|c|c|c|c|c|c|c|c|}
\hline
\textbf{Solver} & \textbf{Domain} & \textbf{Acc. [\%]} & \textbf{steps} & \textbf{t\_max [s]} & \textbf{t\_avg [s]} & \textbf{t\_std [s]}\\
\hline
\hline
Teriyaki-MACRO & MACRO & 95.5 & 10.953 & 54.32 & 8.99 & 4.77\\
\hline
Probe & MACRO & 98.6 & 11.111 & 36.71 & 2.11 & 3.47\\
\hline
\hline
Teriyaki-NO-MACRO & NO-MACRO & 94.0 & 19.158 & 54.71 & 14.61 & 7.16\\
\hline
Probe & NO-MACRO & 98.6 & 22.137 & 43.77 & 2.79 & 3.30\\
\hline
\end{tabular}
\end{center}
\end{table}

As anticipated, Probe is faster than the trained Teriyaki solvers to generate plans end-to-end, in a traditional sense-plan-act workflow.
Yet, Teriyaki solvers still offer decent performance. 
Despite being trained on plans generated by Probe, Teriyaki models are capable of solving problems that Probe failed to process, and even generate \textit{shorter} plans.
The difference in plan length is only $1.5\%$ for the Teriyaki-MACRO, but it raises to $13.5\%$ for Teriyaki-NO-MACRO, which in general leads to plans almost twice as long. 
This seems to suggest that the training procedure rewards shorter completions and that the effect might be stronger the longer the supposed completion gets. 
Nevertheless, this phenomenon requires a more systematic investigation.

\begin{figure}[t!]
\centering
\includegraphics[width=\columnwidth]{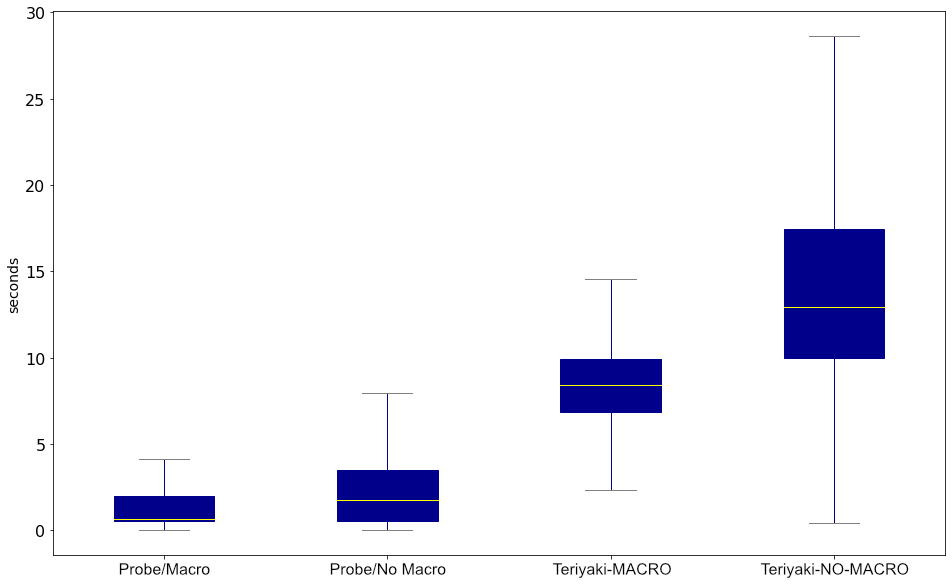}
\caption{
Comparison of the Teriyaki MACRO and NO-MACRO models' planning times against Probe in their respective planning domains.}
\label{fig:planning_times}
\end{figure}

Figure \ref{fig:planning_times} allows for a better assessment of the timing performance of the proposed models, and it compares them to those generated by Probe. 
It can be observed that Teriyaki models do actually scale with the combined length of the input and the output, as hypothesized. 
Planning times of the Teriyaki-NO-MACRO model are approximately twice of the Teriyaki-MACRO model, as expected considering that the plans of the former are approximately twice longer than those of the latter. 
Box plots in the Figure associated with Teriyaki models have a very distinct shape when compared to those of Probe on both domains, which hints at an almost Gaussian distribution of planning times. 
This is coherent with the fact that the plans, which are generated from randomly initialized problems, can assume any length.

\subsection{Action-by-action Plan Streaming}
\label{sec:arch_exp}
One of the major strengths of neurosymbolic action planning using LLMs is that the plan can be streamed as it is being generated.
If properly leveraged, this feature can support action-by-action generation instead of adopting an end-to-end approach.
In many applications involving the use of robots with a required flexible behavior, and especially when frequent re-planning is expected due to changing conditions, simultaneous planning and execution by means of concurrent (sets of) processes could greatly reduce the waiting time for a plan to be available, and therefore executed. 
As we have previously pointed out, this may nicely correlate with a perceived interaction fluency in human-robot collaboration.

In Section \ref{sec:arc_th}, we present an enhanced sense-plan-act architecture that integrates neurosymbolic planning to improve reactivity and resilience to disturbances, while yielding essentially the same plan outputs to traditional approaches. Hence, in this section we focus on a quantitative comparison of the wait times.  

To test the performance of Teriyaki in this regard, we set the Teriyaki-MACRO model to plan against our test dataset until it generated $1000$ actions. As discussed in Section \ref{dataset}, the data that we generated allows for a faithful representation of the real world, while also enabling large-scale testing, without biases induced by the limitations of the robotic platform.

\begin{figure}[t!]
\centering
\includegraphics[width=\columnwidth]{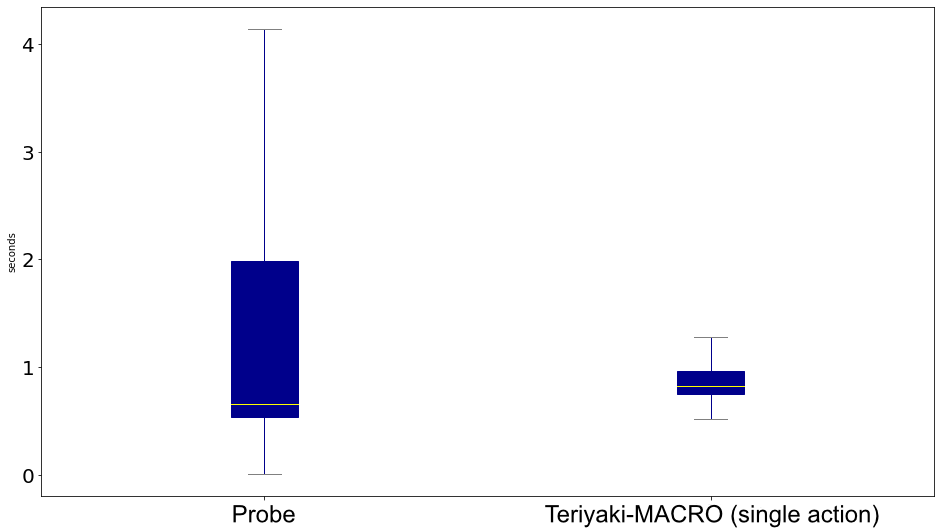}
\caption{
Comparison of the time it takes Teriyaki-MACRO to generate a single action against Probe on the MACRO domain.}
\label{fig:fluency}
\end{figure}

Figure \ref{fig:fluency} compares Teriyaki-MACRO single action timings against Probe timings to generate $1000$ whole plans. 
Not only do we observe that wait times are reduced by 61.2\% on average, but also the response time standard deviation is reduced from 3.47 to just 0.15. This corresponds to a 95.6\% decrease and makes Teriyaki much more predictable than the traditional search-based heuristic solver. 

\subsection{Limitations and Scalability}

We previously pointed out that, although our results are quite promising in so far as they unlock the possibility of simultaneous planning and execution, the data-driven synthesis of a planner as foreseen in the Teriyaki workflow cannot be considered a complete alternative to traditional action planning at this stage.
In fact, a traditional, symbolic action planner is needed to obtain the problem-plan pair data used for training the models and synthesizing the solver.
Herewith, we mention the main current limitations in using the Teriyaki workflow.

\begin{enumerate}
    \item Unlike symbolic planners, our approach is domain-specific and requires training. 
    \item In the current implementation end-to-end planning times are slower than the baseline symbolic planner Probe, \textit{at least} using GPT-3 as base LLM to implement the overall architecture.
    \item Using Transformer-based LLMs, the maximum length of a problem-plan pair is constrained by the context window of the model, which is only $2048$ tokens, or about $8000$ characters, for GPT-3.
    \item Transformer-based LLMs have significant difficulties in generating mathematically accurate results, therefore, while we included advanced PDDL features such as conditional effects, it might be difficult to extend the applicability of this approach to other relevant capabilities such as numerical and temporal planning.
\end{enumerate}

In spite of these limitations, it is noteworthy that there are a number of LLMs, not necessarily based on the Transformer architecture, that are being made available at the writing of this paper and which might help overcome or mitigate such limitations.
For example, newer versions of GPT, such as GPT-3.5-Turbo-16k and GPT-4, extend the context window to $16.000$ and $128.000$ tokens, respectively, and might effectively overcome limitation 3) for all practical planning domains.
Even better, new state space based architectures such as Mamba \citep{gu2023mamba} can be used to implement LLMs completely foregoing the context window, while also presenting greater attention to details in long sequences, and faster inference times.

Apart from utilizing an earlier LLM, further optimizations can enhance Teriyaki's effectiveness. 
Specifically, the planning duration in Teriyaki is affected by the total length of input and output, meaning the lengthiness of action names significantly impacts performance.
As indicated in Figure \ref{fig:planning_times}, implementing a concise action naming convention could align Teriyaki's planning times closer to those seen with Probe.
This underscores the necessity of balancing human readability with efficiency.

Regarding scalability, there are still numerous factors that require further exploration.
Both our study and the work conducted by \cite{pallagani2023plansformer} seem to indicate that sufficiently large models and training datasets, coupled with effective training, enable LLMs to resolve generic planning domains expressed in PDDL syntax, which may include certain advanced elements such as conditional effects.
However, current research has not established a definitive correlation between the intricacy of a planning domain and the requisite model size and training data set needed to consistently address problems within that domain.
The primary challenge lies in the inability, to the best of our knowledge, to quantitatively gauge the complexity of a planning domain.
At most, comparisons between two planning domains can be made based on the dimension of their state space and the quantity of actions inducing state transitions.
Nonetheless, a domain deemed \textit{more complex} by these measures may, paradoxically, be simpler to resolve due to other characteristics of its state space and the specifics of the planner.
For traditional symbolic planners, this is influenced by their heuristic, whereas for LLM-based solvers, it depends on the underlying neural architecture.

In view of the favorable performance scaling discussed in this paper and possibilities offered by simultaneous planning and execution, we consider LLM-based neurosymbolic planning as a promising approach worth considering for future investigation.

\section{Conclusions}

In this paper, we introduce Teriyaki, a framework to generate neurosymbolic, PDDL-compliant planners based on GPT-3. 
In its current implementation, \linebreak Teriyaki relies on a pragmatic and inexpensive procedure for the generation of a training dataset, which is based on the use of an existing action planner able to generate plans on the basis of randomly defined inputs.
The current implementation, the data sets used for training and validation purposes, as well as the ancillary code to generate all the training material from a PDDL-compliant planner, is available open source.
Training can leverage high-performance computing machinery in the cloud, and the resulting model can be deployed to any software architecture for robots adopting standard PDDL-compatible syntax and interfaces. 

The major contribution of the paper is the empirical evidence that this approach can be a viable solution for action planning, as well as a specific proof of concept of the overall workflow. 
In particular, we showed that:
(i) planning validity is on par with that of a traditional, state-of-the-art, PDDL-compliant planner, 
(ii) the average plan length is up to 13.5\% shorter, 
(iii) it is possible to better use Teriyaki in scenarios where robots must re-plan frequently, and in particular due to its ability to generate plans action-by-action, thus reducing average waiting time for a plan by 61.2\%, with standard deviation by 95.6\%. 

Overall, these results hints at a scenario in which the approach can fairly scale up in terms of the number of potentially supported planning domains, even though this remains one of the points to further investigate, and where fluency in human-robot interactions can greatly benefit of the increased reaction times of the robot action planning capabilities.





\section*{Conflict of Interest Statement}

The authors declare that the research was conducted in the absence of any commercial or financial relationships that could be construed as a potential conflict of interest.

\section*{Author Contributions}

AC: Conceptualization, Data curation, Software, Investigation, Writing – original draft. FM: Supervision, Writing – review \& editing. 


\section*{Acknowledgments}
We express heartfelt appreciation to Prof. Mauro Vallati of the University of Huddersfield for his guidance on task planning and for collaborating to the definition of the PDDL domains used in this work.


\section*{Data Availability Statement}
The code and the datasets generated for this study can be found in the following repository: \url{https://github.com/alessiocpt/teriyaki}.

\bibliographystyle{plainnat} 
\bibliography{test}

\begin{thebibliography}{47}
\providecommand{\natexlab}[1]{#1}
\providecommand{\url}[1]{\texttt{#1}}
\expandafter\ifx\csname urlstyle\endcsname\relax
  \providecommand{\doi}[1]{doi: #1}\else
  \providecommand{\doi}{doi: \begingroup \urlstyle{rm}\Url}\fi

\bibitem[Aeronautiques et~al.(1998)Aeronautiques, Howe, Knoblock, McDermott, Ram, Veloso, Weld, SRI, Barrett, Christianson, et~al.]{aeronautiques1998pddl}
Constructions Aeronautiques, Adele Howe, Craig Knoblock, ISI~Drew McDermott, Ashwin Ram, Manuela Veloso, Daniel Weld, David~Wilkins SRI, Anthony Barrett, Dave Christianson, et~al.
\newblock Pddl| the planning domain definition language.
\newblock \emph{Technical Report, Tech. Rep.}, 1998.

\bibitem[Bertolucci et~al.(2019)Bertolucci, Capitanelli, Maratea, Mastrogiovanni, and Vallati]{bertolucci2019automated}
Riccardo Bertolucci, Alessio Capitanelli, Marco Maratea, Fulvio Mastrogiovanni, and Mauro Vallati.
\newblock Automated planning encodings for the manipulation of articulated objects in 3d with gravity.
\newblock In \emph{AI* IA 2019--Advances in Artificial Intelligence: XVIIIth International Conference of the Italian Association for Artificial Intelligence, Rende, Italy, November 19--22, 2019, Proceedings 18}, pages 135--150. Springer, 2019.

\bibitem[Bertolucci et~al.(2021)Bertolucci, Capitanelli, Dodaro, Leone, Maratea, Mastrogiovanni, and Vallati]{bertolucci2021manipulation}
Riccardo Bertolucci, Alessio Capitanelli, Carmine Dodaro, Nicola Leone, Marco Maratea, Fulvio Mastrogiovanni, and Mauro Vallati.
\newblock Manipulation of articulated objects using dual-arm robots via answer set programming.
\newblock \emph{Theory and Practice of Logic Programming}, 21\penalty0 (3):\penalty0 372--401, 2021.

\bibitem[Brown et~al.(2020)Brown, Mann, Ryder, Subbiah, Kaplan, Dhariwal, Neelakantan, Shyam, Sastry, Askell, et~al.]{brown2020language}
Tom Brown, Benjamin Mann, Nick Ryder, Melanie Subbiah, Jared~D Kaplan, Prafulla Dhariwal, Arvind Neelakantan, Pranav Shyam, Girish Sastry, Amanda Askell, et~al.
\newblock Language models are few-shot learners.
\newblock \emph{Advances in neural information processing systems}, 33:\penalty0 1877--1901, 2020.

\bibitem[Capitanelli et~al.(2018)Capitanelli, Maratea, Mastrogiovanni, and Vallati]{capitanelli2018manipulation}
Alessio Capitanelli, Marco Maratea, Fulvio Mastrogiovanni, and Mauro Vallati.
\newblock On the manipulation of articulated objects in human--robot cooperation scenarios.
\newblock \emph{Robotics and Autonomous Systems}, 109:\penalty0 139--155, 2018.

\bibitem[Carf{\'i} et~al.(2019)Carf{\'i}, Foglino, Bruno, and Mastrogiovanni]{Carfietal2019}
Alessandro Carf{\'i}, Francesco Foglino, Barbara Bruno, and Fulvio Mastrogiovanni.
\newblock A multi-sensor dataset for human-human handover.
\newblock \emph{Data in Brief}, 22:\penalty0 119--117, 2019.

\bibitem[Cashmore et~al.(2015)Cashmore, Fox, Long, Magazzeni, Ridder, Carrera, Palomeras, Hurtos, and Carreras]{cashmore2015rosplan}
Michael Cashmore, Maria Fox, Derek Long, Daniele Magazzeni, Bram Ridder, Arnau Carrera, Narcis Palomeras, Natalia Hurtos, and Marc Carreras.
\newblock Rosplan: Planning in the robot operating system.
\newblock In \emph{Proceedings of the international conference on automated planning and scheduling}, volume~25, pages 333--341, 2015.

\bibitem[Chen et~al.(2021)Chen, Tworek, Jun, Yuan, Pinto, Kaplan, Edwards, Burda, Joseph, Brockman, et~al.]{chen2021evaluating}
Mark Chen, Jerry Tworek, Heewoo Jun, Qiming Yuan, Henrique Ponde de~Oliveira Pinto, Jared Kaplan, Harri Edwards, Yuri Burda, Nicholas Joseph, Greg Brockman, et~al.
\newblock Evaluating large language models trained on code.
\newblock \emph{arXiv preprint arXiv:2107.03374}, 2021.

\bibitem[Chen et~al.(2023)Chen, Sun, Liu, Hong, and Gan]{chen2023genome}
Zhenfang Chen, Rui Sun, Wenjun Liu, Yining Hong, and Chuang Gan.
\newblock Genome: Generative neuro-symbolic visual reasoning by growing and reusing modules.
\newblock \emph{arXiv preprint arXiv:2311.04901}, 2023.

\bibitem[Chowdhery et~al.(2022)Chowdhery, Narang, Devlin, Bosma, Mishra, Roberts, Barham, Chung, Sutton, Gehrmann, et~al.]{chowdhery2022palm}
Aakanksha Chowdhery, Sharan Narang, Jacob Devlin, Maarten Bosma, Gaurav Mishra, Adam Roberts, Paul Barham, Hyung~Won Chung, Charles Sutton, Sebastian Gehrmann, et~al.
\newblock Palm: Scaling language modeling with pathways.
\newblock \emph{arXiv preprint arXiv:2204.02311}, 2022.

\bibitem[Dale(2021)]{dale2021gpt}
Robert Dale.
\newblock Gpt-3: What’s it good for?
\newblock \emph{Natural Language Engineering}, 27\penalty0 (1):\penalty0 113--118, 2021.

\bibitem[Darvish et~al.(2018)Darvish, Wanderlingh, Bruno, Simetti, Mastrogiovanni, and Casalino]{Darvishetal2018}
Kourosh Darvish, Francesco Wanderlingh, Barbara Bruno, Enrico Simetti, Fulvio Mastrogiovanni, and Giuseppe Casalino.
\newblock Flexible human-robot cooperation models for assisted shop-floor tasks.
\newblock \emph{Mechatronics}, 51:\penalty0 97--115, 2018.

\bibitem[Darvish et~al.(2021)Darvish, Simetti, Mastrogiovanni, and Casalino]{Darvishetal2021}
Kourosh Darvish, Enrico Simetti, Fulvio Mastrogiovanni, and Giuseppe Casalino.
\newblock A hierarchical architecture for human-robot cooperation processes.
\newblock \emph{IEEE Transactions on Robotics}, 37\penalty0 (2):\penalty0 567--586, 2021.

\bibitem[Driess et~al.(2023)Driess, Xia, Sajjadi, Lynch, Chowdhery, Ichter, Wahid, Tompson, Vuong, Yu, et~al.]{driess2023palm}
Danny Driess, Fei Xia, Mehdi~SM Sajjadi, Corey Lynch, Aakanksha Chowdhery, Brian Ichter, Ayzaan Wahid, Jonathan Tompson, Quan Vuong, Tianhe Yu, et~al.
\newblock Palm-e: An embodied multimodal language model.
\newblock \emph{arXiv preprint arXiv:2303.03378}, 2023.

\bibitem[Garcez and Lamb(2020)]{garcez2020neurosymbolic}
Artur~d'Avila Garcez and Luis~C Lamb.
\newblock Neurosymbolic ai: the 3rd wave.
\newblock \emph{arXiv preprint arXiv:2012.05876}, 2020.

\bibitem[Garrett et~al.(2020)Garrett, Lozano-Perez, and Kaelbing]{Garrettetal2020}
C.~R. Garrett, T.~Lozano-Perez, and L.~P. Kaelbing.
\newblock {PDDLstream}: integrating symbolic planners and blackbox samplers with optimistic adaptive planning.
\newblock In \emph{Proc. 30th International Conference on Automated Planning and Scheduling (ICAPS)}, Anywhere on Earth, October 2020.

\bibitem[Gu and Dao(2023)]{gu2023mamba}
Albert Gu and Tri Dao.
\newblock Mamba: Linear-time sequence modeling with selective state spaces.
\newblock \emph{arXiv preprint arXiv:2312.00752}, 2023.

\bibitem[Hatcher and Yu(2018)]{hatcher2018survey}
William~Grant Hatcher and Wei Yu.
\newblock A survey of deep learning: Platforms, applications and emerging research trends.
\newblock \emph{IEEE Access}, 6:\penalty0 24411--24432, 2018.

\bibitem[Helmert(2006)]{helmert2006fast}
Malte Helmert.
\newblock The fast downward planning system.
\newblock \emph{Journal of Artificial Intelligence Research}, 26:\penalty0 191--246, 2006.

\bibitem[Heyer(2010)]{heyer2010human}
Clint Heyer.
\newblock Human-robot interaction and future industrial robotics applications.
\newblock In \emph{2010 ieee/rsj international conference on intelligent robots and systems}, pages 4749--4754. IEEE, 2010.

\bibitem[Hoffman(2019)]{hoffman2019evaluating}
Guy Hoffman.
\newblock Evaluating fluency in human--robot collaboration.
\newblock \emph{IEEE Transactions on Human-Machine Systems}, 49\penalty0 (3):\penalty0 209--218, 2019.

\bibitem[Howey et~al.(2004)Howey, Long, and Fox]{howey2004val}
Richard Howey, Derek Long, and Maria Fox.
\newblock Val: Automatic plan validation, continuous effects and mixed initiative planning using pddl.
\newblock In \emph{16th IEEE International Conference on Tools with Artificial Intelligence}, pages 294--301. IEEE, 2004.

\bibitem[Huang et~al.(2022)Huang, Abbeel, Pathak, and Mordatch]{huang2022language}
Wenlong Huang, Pieter Abbeel, Deepak Pathak, and Igor Mordatch.
\newblock Language models as zero-shot planners: Extracting actionable knowledge for embodied agents.
\newblock In \emph{International Conference on Machine Learning}, pages 9118--9147. PMLR, 2022.

\bibitem[Lipovetzky and Geffner(2011)]{lipovetzky2011searching}
Nir Lipovetzky and Hector Geffner.
\newblock Searching for plans with carefully designed probes.
\newblock In \emph{Proceedings of the International Conference on Automated Planning and Scheduling}, volume~21, pages 154--161, 2011.

\bibitem[Logeswaran et~al.(2022)Logeswaran, Fu, Lee, and Lee]{logeswaran2022few}
Lajanugen Logeswaran, Yao Fu, Moontae Lee, and Honglak Lee.
\newblock Few-shot subgoal planning with language models.
\newblock \emph{arXiv preprint arXiv:2205.14288}, 2022.

\bibitem[Loshchilov and Hutter(2016)]{loshchilov2016sgdr}
Ilya Loshchilov and Frank Hutter.
\newblock Sgdr: Stochastic gradient descent with warm restarts.
\newblock \emph{arXiv preprint arXiv:1608.03983}, 2016.

\bibitem[Macci{\'o} et~al.(2022)Macci{\'o}, Carf{\'i}, and Mastrogiovanni]{Maccioetal2022}
Simone Macci{\'o}, Alessandro Carf{\'i}, and Fulvio Mastrogiovanni.
\newblock A system for hierarchical planning in service mobile robotics.
\newblock In \emph{2022 IEEE International Conference on Robotics and Automation}, 2022.

\bibitem[Mastrogiovanni et~al.(2004)Mastrogiovanni, Sgorbissa, and Zaccaria]{Mastrogiovannietal2004}
Fulvio Mastrogiovanni, Antonio Sgorbissa, and Renato Zaccaria.
\newblock A system for hierarchical planning in service mobile robotics.
\newblock In \emph{8th International Conference on Intelligent Autonomous Systems}, 2004.

\bibitem[Mikolov et~al.(2011)Mikolov, Kombrink, Burget, {\v{C}}ernock{\`y}, and Khudanpur]{mikolov2011extensions}
Tom{\'a}{\v{s}} Mikolov, Stefan Kombrink, Luk{\'a}{\v{s}} Burget, Jan {\v{C}}ernock{\`y}, and Sanjeev Khudanpur.
\newblock Extensions of recurrent neural network language model.
\newblock In \emph{2011 IEEE international conference on acoustics, speech and signal processing (ICASSP)}, pages 5528--5531. IEEE, 2011.

\bibitem[Murali et~al.(2020)Murali, Darvish, and Mastrogiovanni]{Muralietal2020}
Prajval~Kumar Murali, Kourosh Darvish, and Fulvio Mastrogiovanni.
\newblock Deployment and evaluation of a flexible human-robot collaboration model based on and/or graphs in a manufacturing environment.
\newblock \emph{Intelligent Service Robotics}, 13:\penalty0 439--457, 2020.

\bibitem[Oussidi and Elhassouny(2018)]{oussidi2018deep}
Achraf Oussidi and Azeddine Elhassouny.
\newblock Deep generative models: Survey.
\newblock In \emph{2018 International conference on intelligent systems and computer vision (ISCV)}, pages 1--8. IEEE, 2018.

\bibitem[Pallagani et~al.(2023)Pallagani, Muppasani, Srivastava, Rossi, Horesh, Murugesan, Loreggia, Fabiano, Joseph, Kethepalli, et~al.]{pallagani2023plansformer}
Vishal Pallagani, Bharath Muppasani, Biplav Srivastava, Francesca Rossi, Lior Horesh, Keerthiram Murugesan, Andrea Loreggia, Francesco Fabiano, Rony Joseph, Yathin Kethepalli, et~al.
\newblock Plansformer tool: Demonstrating generation of symbolic plans using transformers.
\newblock In \emph{IJCAI}, volume 2023, pages 7158--7162. International Joint Conferences on Artificial Intelligence, 2023.

\bibitem[Scao et~al.(2022)Scao, Fan, Akiki, Pavlick, Ili{\'c}, Hesslow, Castagn{\'e}, Luccioni, Yvon, Gall{\'e}, et~al.]{scao2022bloom}
Teven~Le Scao, Angela Fan, Christopher Akiki, Ellie Pavlick, Suzana Ili{\'c}, Daniel Hesslow, Roman Castagn{\'e}, Alexandra~Sasha Luccioni, Fran{\c{c}}ois Yvon, Matthias Gall{\'e}, et~al.
\newblock Bloom: A 176b-parameter open-access multilingual language model.
\newblock \emph{arXiv preprint arXiv:2211.05100}, 2022.

\bibitem[Silver et~al.(2022)Silver, Hariprasad, Shuttleworth, Kumar, Lozano-P{\'e}rez, and Kaelbling]{silver2022pddl}
Tom Silver, Varun Hariprasad, Reece~S Shuttleworth, Nishanth Kumar, Tom{\'a}s Lozano-P{\'e}rez, and Leslie~Pack Kaelbling.
\newblock Pddl planning with pretrained large language models.
\newblock In \emph{NeurIPS 2022 Foundation Models for Decision Making Workshop}, 2022.

\bibitem[Silver et~al.(2023)Silver, Dan, Srinivas, Tenenbaum, Kaelbling, and Katz]{silver2023generalized}
Tom Silver, Soham Dan, Kavitha Srinivas, Joshua~B Tenenbaum, Leslie~Pack Kaelbling, and Michael Katz.
\newblock Generalized planning in pddl domains with pretrained large language models.
\newblock \emph{arXiv preprint arXiv:2305.11014}, 2023.

\bibitem[Singh et~al.(2023)Singh, Blukis, Mousavian, Goyal, Xu, Tremblay, Fox, Thomason, and Garg]{singh2023progprompt}
Ishika Singh, Valts Blukis, Arsalan Mousavian, Ankit Goyal, Danfei Xu, Jonathan Tremblay, Dieter Fox, Jesse Thomason, and Animesh Garg.
\newblock Progprompt: Generating situated robot task plans using large language models.
\newblock In \emph{2023 IEEE International Conference on Robotics and Automation (ICRA)}, pages 11523--11530. IEEE, 2023.

\bibitem[Smith et~al.(2022)Smith, Patwary, Norick, LeGresley, Rajbhandari, Casper, Liu, Prabhumoye, Zerveas, Korthikanti, et~al.]{smith2022using}
Shaden Smith, Mostofa Patwary, Brandon Norick, Patrick LeGresley, Samyam Rajbhandari, Jared Casper, Zhun Liu, Shrimai Prabhumoye, George Zerveas, Vijay Korthikanti, et~al.
\newblock Using deepspeed and megatron to train megatron-turing nlg 530b, a large-scale generative language model.
\newblock \emph{arXiv preprint arXiv:2201.11990}, 2022.

\bibitem[Song et~al.(2023)Song, Wu, Washington, Sadler, Chao, and Su]{song2023llm}
Chan~Hee Song, Jiaman Wu, Clayton Washington, Brian~M Sadler, Wei-Lun Chao, and Yu~Su.
\newblock Llm-planner: Few-shot grounded planning for embodied agents with large language models.
\newblock In \emph{Proceedings of the IEEE/CVF International Conference on Computer Vision}, pages 2998--3009, 2023.

\bibitem[Sundermeyer et~al.(2012)Sundermeyer, Schl{\"u}ter, and Ney]{sundermeyer2012lstm}
Martin Sundermeyer, Ralf Schl{\"u}ter, and Hermann Ney.
\newblock Lstm neural networks for language modeling.
\newblock In \emph{Thirteenth annual conference of the international speech communication association}, 2012.

\bibitem[Sussman(1973)]{sussman1973computational}
Gerald~J Sussman.
\newblock A computational model of skill acquisition.
\newblock 1973.

\bibitem[Thoppilan et~al.(2022)Thoppilan, De~Freitas, Hall, Shazeer, Kulshreshtha, Cheng, Jin, Bos, Baker, Du, et~al.]{thoppilan2022lamda}
Romal Thoppilan, Daniel De~Freitas, Jamie Hall, Noam Shazeer, Apoorv Kulshreshtha, Heng-Tze Cheng, Alicia Jin, Taylor Bos, Leslie Baker, Yu~Du, et~al.
\newblock Lamda: Language models for dialog applications.
\newblock \emph{arXiv preprint arXiv:2201.08239}, 2022.

\bibitem[Touvron et~al.(2023)Touvron, Martin, Stone, Albert, Almahairi, Babaei, Bashlykov, Batra, Bhargava, Bhosale, et~al.]{touvron2023llama}
Hugo Touvron, Louis Martin, Kevin Stone, Peter Albert, Amjad Almahairi, Yasmine Babaei, Nikolay Bashlykov, Soumya Batra, Prajjwal Bhargava, Shruti Bhosale, et~al.
\newblock Llama 2: Open foundation and fine-tuned chat models.
\newblock \emph{arXiv preprint arXiv:2307.09288}, 2023.

\bibitem[Valmeekam et~al.(2022)Valmeekam, Olmo, Sreedharan, and Kambhampati]{valmeekam2022large}
Karthik Valmeekam, Alberto Olmo, Sarath Sreedharan, and Subbarao Kambhampati.
\newblock Large language models still can't plan (a benchmark for llms on planning and reasoning about change).
\newblock \emph{arXiv preprint arXiv:2206.10498}, 2022.

\bibitem[Vaswani et~al.(2017)Vaswani, Shazeer, Parmar, Uszkoreit, Jones, Gomez, Kaiser, and Polosukhin]{vaswani2017attention}
Ashish Vaswani, Noam Shazeer, Niki Parmar, Jakob Uszkoreit, Llion Jones, Aidan~N Gomez, {\L}ukasz Kaiser, and Illia Polosukhin.
\newblock Attention is all you need.
\newblock \emph{Advances in neural information processing systems}, 30, 2017.

\bibitem[Wake et~al.(2023)Wake, Kanehira, Sasabuchi, Takamatsu, and Ikeuchi]{wake2023chatgpt}
Naoki Wake, Atsushi Kanehira, Kazuhiro Sasabuchi, Jun Takamatsu, and Katsushi Ikeuchi.
\newblock Chatgpt empowered long-step robot control in various environments: A case application.
\newblock \emph{arXiv preprint arXiv:2304.03893}, 2023.

\bibitem[Wang et~al.(2023)Wang, Wang, Wang, Cao, Saurous, and Kim]{wang2023grammar}
Bailin Wang, Zi~Wang, Xuezhi Wang, Yuan Cao, Rif~A Saurous, and Yoon Kim.
\newblock Grammar prompting for domain-specific language generation with large language models.
\newblock \emph{arXiv preprint arXiv:2305.19234}, 2023.

\bibitem[Wang et~al.(2021)Wang, Wang, Joty, and Hoi]{wang2021codet5}
Yue Wang, Weishi Wang, Shafiq Joty, and Steven~CH Hoi.
\newblock Codet5: Identifier-aware unified pre-trained encoder-decoder models for code understanding and generation.
\newblock \emph{arXiv preprint arXiv:2109.00859}, 2021.

\end{thebibliography}

\end{document}